\documentclass[11pt, a4paper, twoside]{article}
\usepackage[margin=4.0cm]{geometry}
\usepackage{epsfig}
\usepackage{color}
\usepackage{pictex}
\usepackage{epsf}
\usepackage[small]{caption}

\newcommand{\psfigure}[3]{
        \begin{figure}[h!]
        \begin{center}
        \epsfxsize=#2 \epsfbox{#1.eps}
        \end{center}
	\vspace{-0.2in} 
        \caption{#3}
        \label{fig:#1}
        \end{figure}
}
\newcommand{\psfigureab}[5]{
        \begin{figure}[h!]
        \begin{center}
	\begin{tabular}{cc}
	        \epsfxsize=#4 \epsfbox{#2.eps} &
        	\epsfxsize=#4 \epsfbox{#3.eps} \\
		(a) & (b) \\
	\end{tabular}
        \end{center}
        \vspace{-0.2in}
        \caption{#5}
        \label{fig:#1}
        \end{figure}
}

\newcommand{\psfigureabcd}[7]{
        \begin{figure}[h!]
        \smallskip
        \begin{center}
	\begin{tabular}{cc}
	        \epsfxsize=#6 \epsfbox{#2.eps} &
        	\epsfxsize=#6 \epsfbox{#3.eps} \\
		(a) & (b) \\[14pt]
	        \epsfxsize=#6 \epsfbox{#4.eps} &
        	\epsfxsize=#6 \epsfbox{#5.eps} \\
		(c) & (d) \\
	\end{tabular}
        \end{center}
        \caption{#7}
        \label{fig:#1}
        \end{figure}
}

\begin{document}
\begin{center}
{\Large
	An effective edge--directed
	frequency filter for removal of aliasing in
	upsampled images \\
}

\vspace{0.1in}

{\small
	Artur Rataj, Institute of Theoretical and Applied
	Computer Science, Ba\l tycka 5, Gliwice, Poland \\
}
\end{center}

\vspace{0.1in}

{\small
        {\bf Abstract. } Raster images can
        have a range of various distortions
	connected to their raster structure.
        Upsampling them might in effect substantially yield
        the raster structure of the original image, known as
        aliasing. The upsampling itself may introduce aliasing
        into the upsampled image as well.
	The presented method attempts to remove the
        aliasing using frequency filters based on the discrete fast Fourier
        transform, and applied directionally in certain regions placed along
	the edges in the image.

	As opposed to some anisotropic smoothing methods, the presented
	algorithm aims to selectively reduce only the aliasing, preserving
	the sharpness of image details.

	The method can be used as a post--processing filter along with various
	upsampling algorithms. It was experimentally shown that the method can
	improve the visual quality of the upsampled images.

	{\textbf{keywords:}
		aliasing, upsampling, frequency filter, edge detection
	}
}

\section{Introduction}
\label{sec:introduction}
    Raster images often have distortions connected with their raster structure.
    These distortions can for example be an undersampling, distorted
    intensity response curves or processing like sharpening or unsharp mask.
    Upsampling the distorted images, using for example
    the bicubic interpolation \cite{keys1981bicubic, mitchell1988reconstruction},
    might in effect substantially yield the raster structure
    of the original image, what is known in image processing
    as aliasing \cite{mitchell1988reconstruction}.
    Additionally, upsampling methods that attempt to produce sharp images,
    might have an intrinsic trait of introducing the aliasing \cite{
    mitchell1988reconstruction}. The
    presented method attempts to remove the aliasing artifacts
    using frequency filters based on the discrete fast Fourier transform,
    and applied directionally
    in certain regions placed along the edges detected in the image. The selective
    directional applying of these filters
    serves the purpose of estimating the presence of
    the aliasing in the places where it is likely to occur, and where it
    is at the same time
    unlikely that the objects in the image will be confused with the
    aliasing.

    The special feature of the method is that it aims to selectively
    reduce the aliasing,
    trying at the same time to preserve the sharpness of image details.
    It makes it different from typically used interpolations like
    the bilinear or bicubic ones \cite{keys1981bicubic, mitchell1988reconstruction},
    that
    produce images that are blurry or aliased, or
    various anisotropic smoothing methods like these described in
    \cite{tschumperle2005regularization, tschumperle2006anisotropic},
    that aim to generally
    smoothen objects in the image, what might lead, as it will be illustrated
    in tests, to very unnatural looking images.
    On of the more widely used complex image restoration methods -- 
    NEDI \cite{xinli2001interpolation}, also makes some
    textures look unnatural and still produces substantial
    aliasing in some images.

    The following sections discuss, in order, aliasing,
    a custom sub--pixel precision edge detection method used to
    direct the filtering,
    and the frequency filtering. Finally, some tests are presented.

\section{Aliasing}
\begin{figure}
\begin{center}
\begin{tabular}{cccc}
\includegraphics[width=0.4in]{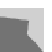} & \includegraphics[width=0.15in]{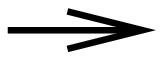} & \includegraphics[width=0.4in]{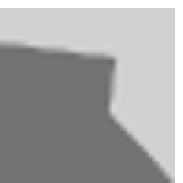} \\
\includegraphics[width=0.4in]{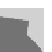} & \includegraphics[width=0.15in]{arrow.eps} & \includegraphics[width=0.4in]{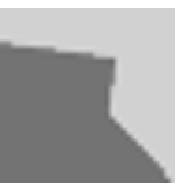}\\
\includegraphics[width=0.4in]{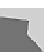} & \includegraphics[width=0.15in]{arrow.eps} & \includegraphics[width=0.4in]{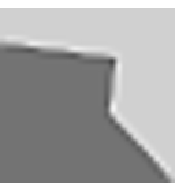}\\
\end{tabular}
\vspace{-0.1in} 
\caption{An example of aliasing. The first column contains
original \(64 \times 64\) images, that are, in subsequent rows,
with only small distortions, distorted gray response
curves and sharpened. The second column contains corresponding
\(128 \times 128\) upsampled images, using the bicubic interpolation.}
\label{fig:raster-artifacts-example}
\end{center}
\end{figure}

The discussed aliasing in the upsampled images is connected with
the raster of the source image, and not of the upsampled image. In
Fig.~\ref{fig:raster-artifacts}, a schematic example of an object
upsampled four
times in each direction is shown. Bold lines show borders of the
original pixels, smallest rectangles show borders of the pixels in the
upsampled image.
\psfigure{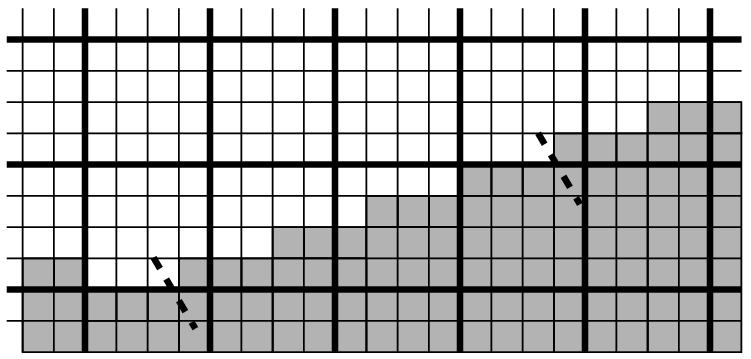}
{1.4in}
{A schematic example of upsampling.
}
The image shows a dark object on a white background.
The object boundary in
the original image consisted of pixels whose brightness changed
approximately periodically, with the period connected to the period of
passing of the horizontal line between
the pixels in the original raster.
It can be seen in the upsampled image -- the
brighter boundary pixels in the original image have corresponding \(4
\times 4\) pixel blocks in the upsampled image that consist of mostly white
pixels, and conversely, the darker pixels in the original image have
corresponding blocks of mostly dark pixels.
Similarly, of course, if boundary would be more close to a vertical one,
the period of passing of the vertical raster lines would be important in turn.
As can be seen in the
example in Fig.~\ref{fig:raster-artifacts-example}, various distortions
of the image
may cause `waving' of location,
color or sharpness of the upsampled boundaries, depending on the
particular distortion and the upscaling method. What is important here,
though, is
that the period \(l_{0}\) of the `waving' for a straight boundary
is the same as the
period of the brightness variability of the pixels in the original image,
which in turn, as it was discussed and also can be
seen in Fig.~\ref{fig:raster-artifacts-example}, is approximately
equal to the
length of the object border between two either horizontal or vertical
subsequent lines of the original raster, depending on
if the border is either more close to, respectively, the horizontal or
the vertical direction. For a straight border, \(l_{0}\) is thus as follows:
\begin{equation}
	l_{0} = \left\{
		\begin{array}{ll}\displaystyle
		U \left|\frac{x_{l} - x_{0}}{y_{l} - y_{0}}\right| &
			\textrm{if \(|x_{l} - x_{0}| \ge |y_{l} - y_{0}|\)} \\[8pt]
		\displaystyle
		U \left|\frac{y_{l} - y_{0}}{x_{l} - x_{0}}\right| &
			\textrm{if \(|x_{l} - x_{0}| < |y_{l} - y_{0}|\)} \\
		\end{array}
	\right.
\end{equation}
where \(U\) is the scale of the upsampling,
\((x_{0}, y_{0})\) is the first pixel of a straight
fragment of a boundary and
\((x_{l}, y_{l})\) is the last pixel of the fragment,
using the coordinates of the upsampled image.
If the fragment is only approximately straight, the
equation gives an approximate common \(l_{0}\),
while local periods can vary along the fragment.
An example of such an approximately straight fragment is
illustrated in Fig.~\ref{fig:frequency-filtering}.

Thus, estimation the period on basis
of the orientation of a border might be a good way of
detecting the corresponding artifacts, what in turn might be the first stage
of reducing these detected artifacts. This is the basic presume of the presented
method. 

\section{Sub--pixel precision edge detection}
    Edge detection \cite{marr1980edge, canny1986edge, jain1989fundamentals,
    ziou1998over} in raster images is one of the basic methods of
    feature extraction from images. This paper employs a
    simple low--level definition of an edge described in
    \cite{martin2002learning}: an abrupt change in some low--level image
    feature as brightness or color, as opposed to a boundary, described in
    the cited paper as a higher--level feature.  The presented edge detection
    method is designed to give edges with sub--pixel precision,
    and to detect even small discontinuities in the image. This is because
    the aim is, as opposed to typical edge detection methods,
    not to extract the more prominent edges, but to get a precise edge
    map for frequency filtering. Additionally, the edge detector
    employed must have a high 
    resistance to the image distortions discussed, like undersampling.
    This is why a custom edge detector was designed.

\subsection{Finding edges}
\label{sec:edge_detection}
        In the first step of the edge detection, a Sobel
        operator \cite{sobel1970camera}
        is applied to the upsampled image. If the image has multiple bands,
        each one is processed separately and then
        the resulting images are averaged into one
        single--band image.
        Then, the roof edges \cite{perona1991detecting, baker1998parametric}
        are searched for in that resulting image.
        As the Sobel operator produces a gradient map--like image,
        the roof edges obtained are effectively the discontinuity
        edges as discussed in \cite{martin2002learning}.
	
	To detect the roof edges, an operator called
	peakiness detection is used.
    The edge detection basically works by finding `bumps' in the gradient image, at
    various angles. The computational complexity is kept low by
    using the following approach.
    For each of the angles
    \(a_{i} = (i + 0.5)\frac{1}{2}\pi/N\), \(i = 0,\,\ldots\,N - 1\), scan the image
    along lines that are at the angle \(a_{i}\) to the horizontal axis, so that:
    \begin{itemize}
        \item if \(a_{1} \le \pi/4\), let the consecutive lines be one vertical
        pixel apart;
        \item if \(a_{1} > \pi/4\), let the consecutive lines be one horizontal
        pixel apart;
    \end{itemize}
    \psfigure{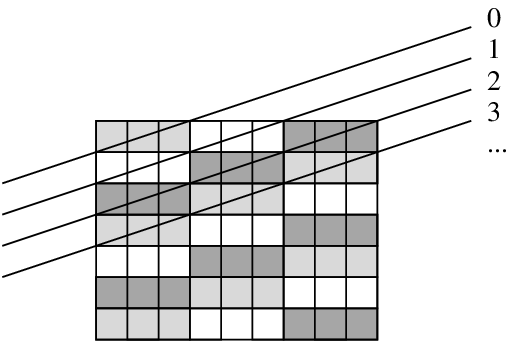}
            {1.4in}
            {A series of scan lines for a given angle.}
    and let the lines cover such a range, that, together, they cover the entire area
    of the image. An example case for \(a_{1} \le \pi/4\)
    is illustrated in Fig.~\ref{fig:scan_lines}. As it can be seen,
    such a way of aligning subsequent lines provides that all pixels
    are covered, in scans for each \(a_{1}\). Yet, there is not a separate searching
    for `bumps' around each pixel at an angle \(a_{i}\) --
    sequential searching for `bumps' on a single line at an angle \(a_{i}\)
    covers searching for `bumps' for each pixel on that line, what decreases
    the mentioned computational complexity. The value of \(N = 7\) was
    chosen, as a precise enough and making the scanning reasonably
    fast at the same time.

    The `bump' criterion is as follows.
    Let \(p_{0},\,p_{1},\,\ldots\,p_{M - 1}\) be intensities of subsequent pixels
    on a given scanned line of \(M\) pixels.
    The searching for `bumps' within a single
    line works as follows:  
    for the pixel \(n\)th, if its intensity is larger by \(d\) than both the
    intensity of the pixel \((n - r)\)th and the intensity of the pixel
    \((n + r)\)th, then increase
    the `peakiness' of the pixel by 1. The coefficients \(d\) and \(r\) should be
    large enough to reduce single--pixel level noise, and small enough
    to maintain good edge location.
    To improve the detection of edges at various scales,
    \(p_{\max} = 3\) passes of the edge detection are
    performed, each modifying common `peakiness' of a pixel,
    with three different sets of values for \(d\) and \(r\):
    \begin{equation}
	\begin{array}{c}
		p = 1, 2, \ldots p_{\max} \\
		r_{p} = p + 2 \\
                d_{p} = 0.015 + 0.005p \\
	\end{array}
    \end{equation}
    where the index \(p\) denotes a respective pass. If the image processed
    is very blurry, \(r\) might require an appropriate increase.

    Because 
    \(p_{\max}N > 1\) scanning lines pass through each pixel,
    one for each angle, the `peakiness' is an averaged
    value of several tests for the `bumps',
    what may obviously reduce single--pixel level noise.

    Only pixels whose accumulated peakiness value is equal or larger
    than a given
    threshold \(e_{\min}\) are regarded as the edge ones,
    to reduce the detection of what is an image noise, and not a real edge.
    The value of \(e_{\min} = 6\) was adjusted in tests. It can be decreased
    for images with low noise and weak edges, and increased for images with
    high level of noise.

    The roof edges obtained using this method
    can be thick, while the needed edges must be one--pixel
    wide. To correct that, centers of the roof edges are extracted using
    a simple thinning method, for example that described in
    \cite{gonzales1993image}, with the 8--neighborhood criterion. 

\subsection{Correction of the edges}
\label{sec:undersampling}
   The discussed distortions, the edge detection method itself, or
   image noise may decrease the quality of the obtained edges.
   Therefore, cleaning of the edges from small branches
   and protruding pixels, and the reduction of `waving' of the
   edges, is used.

\subsubsection{Cleaning the edges}
\label{sec:cleaning}
Both the method of the waving reduction, and the finding of
approximately straight fragments discussed later in
Sec.~\ref{sec:frequency-filtering}, are sensitive to two kinds of
`noise' of the edges -- small branches and single protruding
pixels. Example of such distortions is shown in
Fig.~\ref{fig:edge-distortions}. The work--around is straightforward --
edges below a given length are deleted, where each pixel
connecting
three or more branches is considered a boundary between the edges.
In the first iteration,
edges of the length of \(1\) are deleted, then edges of the length \(2\), and
so on, till some value \(L_{\min} - 1\), with re--measuring of the edge lengths
after each iteration. If, instead, we'd immediately
begin with deleting all edges of length less or equal than \(L_{\min}\),
then the edges like the grayed one in Fig.~\ref{fig:edge-distortions}
would be deleted, instead of only the two small branches visible
in the image.

The small protruding single pixels, like that seen in
Fig.~\ref{fig:edge-distortions}, are moved
back to the edge, using a trivial method.
\psfigure{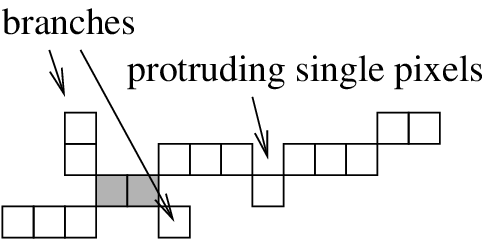}
	{1.4in}
	{Example edge `noise' to be cleaned.}

\subsubsection{Reduction of waving}
\label{sec:reduction-of-waving}
\begin{figure}
\begin{center}
\begin{tabular}{ccccc}
\includegraphics[width=0.4in]{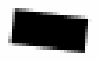} & \includegraphics[width=0.15in]{arrow.eps} & \includegraphics[width=0.4in]{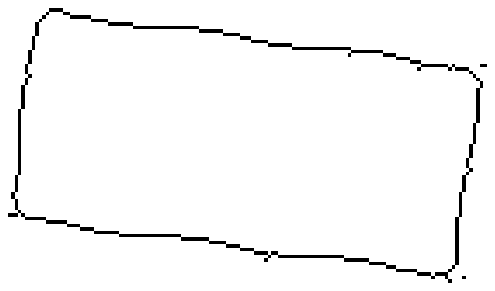} & \includegraphics[width=0.15in]{arrow.eps} & \includegraphics[width=0.4in]{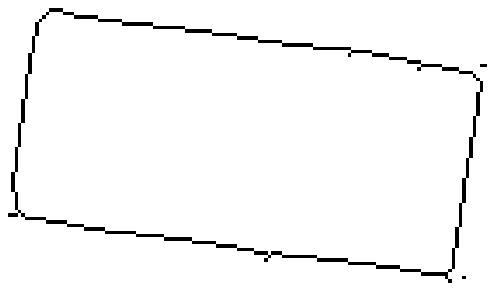} \\[-5pt]
\includegraphics[width=0.4in]{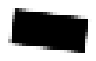} & \includegraphics[width=0.15in]{arrow.eps} & \includegraphics[width=0.4in]{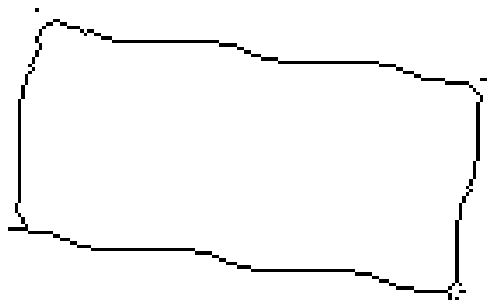} & \includegraphics[width=0.15in]{arrow.eps} & \includegraphics[width=0.4in]{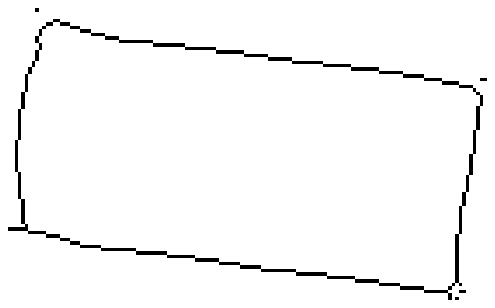} \\
\end{tabular}
\vspace{-0.1in}
\caption{An example of the reduction of waving for an image with two
different gray response curves.}
\label{fig:undersampling}
\end{center}
\end{figure}

    Aliasing in the upsampled image may produce variously `waving' edges,
    An example of `waving', and its correction,
    is shown in Fig.~\ref{fig:undersampling}.
    The edge detector should be resistant to the aliasing artifacts,
    thus, the reduction of the waving is performed.
The procedure to reduce waving has the following steps:

\begin{enumerate}\addtolength{\itemsep}{-0.3\baselineskip}
\item Find junctions, that is corners of pixels that have two neighboring
edge pixels.

\item For each of these two pixels, find the length of rectangular
sequences \(S\) that
begin at the subsequent edge pixel. A rectangular sequence is a sequence of
4--neighboring pixels that is either vertical or horizontal.

\item If one of these cases occurs: both rectangular sequences \(S\) are
horizontal, or both are vertical, one consists of a single pixel
and the other has more than one pixel, then such a junction can
be classified as, accordingly, either horizontal or vertical.
In such a case, then, it is assumed that the junction is a part of
some edge \(E\) that can, respectively, be classified as being locally
either closer to some horizontal or vertical direction.

\item if \(E\) could be classified as locally closer to horizontal or vertical
direction, the junction is marked as a movable one along that closer direction,
that is, able to modify \(E\), by shortening the
longer \(S\) and extending the shorter \(S\), as it is shown in the example
in Fig.~\ref{fig:junctions}.
	\psfigureab{junctions}
		{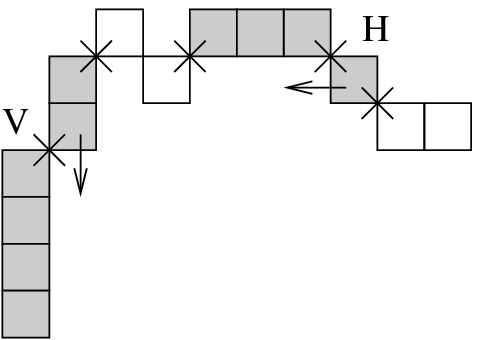}
		{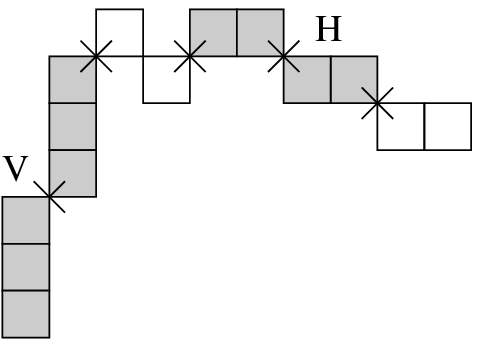}
		{0.9in}
		{(a) Examples of junctions.
		 The edge pixels are marked with rectangles.
		 All junctions are marked with crosses.
		 Only two junctions, for clarity, have their
		 sequences \(S\) marked with gray color.
		 Possible moving direction of these two junctions
		 are marked with arrows. The junction \(H\) is the
		 horizontal one, and the junction \(V\) is the 
		 vertical one.
		 (b) The same edge, the junctions \(H\) and \(V\)
		 were moved
		 by the length of one pixel.}

\item For each movable junction,
set the maximum length \(l_{\max}\) the junction is
allowed to move. This constraint exists to prevent the edges from too
large moves. Let the two rectangular sequences \(S\)
neighboring to a junction
have their respective lengths \(s_{1}\) and
\(s_{2}\). Then,
\begin{equation}
\label{eq:junction-move}
l_{\max} = \min\big[\max\left(s_{1}, s_{2}\right),\,
	l_{1}\min\left(s_{1}, s_{2}\right) + l_{2}\big]
\end{equation}
The basic limitation in (\ref{eq:junction-move})
is that \(l_{\max} \le \max\left(s_{1}, s_{2}\right)\),
thus,
an approximately straight edge can not be moved aside by more than about one
pixel. The coefficients \(l_{1}\) and \(l_{2}\) precisely regulate
\(l_{\max}\). It was found in tests, that \(l_{1} = 3\) and \(l_{2} = 1\)
gives a good trade--off between effective waving reduction and
a none to moderate displacement of the edges.

\item
After \(l_{\max}\) was determined for each movable junction,
the following sub--procedure \(W\) is repeatedly performed,
each time for the whole image,
until either the number of the repetitions of performing
 \(W\) reaches a given number \(N_{w} = 50\),
or the stability is reached, that is a given run of
\(W\) does not change anything in the image.
The limitation by using \(N_{w}\) is only to prevent the wave reduction
from taking too long time.

\(W\) is as follows.
For each movable junction, if \(|s_{1} - s_{2}| > 1\), and the junction did not reach
its \(l_{\max}\) value, move the junction so that to shorten its larger
sequence \(S\) by one pixel and extend its shorter sequence \(S\) by one pixel.
The condition \(|s_{1} - s_{2}| > 1\) exists to cause the rectangular
sequences \(S\)
to have more similar length, with the prevention of a junction to be moved
back and forth in subsequent executions of the procedure \(W\).
\end{enumerate}

\section{Frequency filtering}
\label{sec:frequency-filtering}
The frequency filtering has two stages: in the first stage, find approximately straight
fragments of edges, and in the second stage,
do frequency filtering directed along each
such a fragment. Because the fragments are approximately straight, a common
approximate base period of aliasing artifacts can be determined for
each fragment. Such
a base period \(l_{0}\) is used then
in filtering the frequency spectrums.

\subsection{Finding approximately straight fragments}
\label{sec:approximately-straight-fragments}
An approximately straight fragment \(F\)
is an edge or a part of the edge. A fragment \(F\) can not include the
branch pixels, that is those that have more than two neighbors being
edge pixels.
The criterion of a fragment to be approximately straight is very simple:
all pixels of the fragment must not be further from the straight line between
two endings of \(F\) by more than \(d = s_{d}U\). \(U\)
is the scale of upsampling,
and it occurs in the formula because image features
are linearly proportional to \(U\), and \(s_{d}\) is a coefficient regulating
how approximately straight \(F\) should be. It was determined
in tests that the value of \(s_{d} = 0.4\) is a good trade--off between
many short fragments for small \(s_{d}\) and bad approximation of
common \(l_{0}\) for the whole \(F\) for large \(s_{d}\).
There is yet another condition, \(Q\), for \(F\): its subsequent
pixels must all either have always increasing or always decreasing
\(x\) coordinates or \(y\) coordinates. If the condition applies
to the \(x\) coordinates, the fragment \(F\) is called a horizontal one,
and otherwise it is called the vertical one. An example of a horizontal 
\(F\) is shown in Fig.\ref{fig:frequency-filtering}. The need for the
condition \(Q\) will be explained in Sec.~\ref{sec:fragment-directed}.

The fragments \(F\) are searched for as follows:
find an edge pixel \(P\) that is
a part of an edge whose pixels are not assigned to
any fragment \(F\) yet. Trace the unassigned edge pixels from the pixel \(P\),
using the 8--neighborhood criterion, 
the same that was used during thinning of the edges.
Do that until
the end of the unassigned edge pixels is found, or a branch pixel is
found. Then trace the pixels back to
search for the other end of these unassigned pixels,
until the other end is reached or
the criterion of approximate straight edge stops to be fulfilled,
and this way find a new \(F\).  
Then set the pixels of the new \(F\) as ones
assigned to \(F\), and continue searching for fragments \(F\)
until all pixels, excluding the
branch pixels, are assigned to some \(F\).

\subsection{Fragment-directed frequency filtering}
\label{sec:fragment-directed}
It is important for the frequency filtering to be applied possibly
along object boundaries. Applying it, for example, across fence pales,
may alter important image matter, as the periodicity of
occurring of the pales might be confused
with aliasing. This is why the edge detector is employed,
and then the fragments \(F\) are extracted.  

With each \(F\), the filtering strength \(S_{f}\) is estimated. \(S_{f}\)
is directly related to the size of the region along \(F\) that is filtered,
as illustrated in Fig.~\ref{fig:frequency-filtering} -- the fragment
is moved \(S_{f}\) times up and \(S_{f}\) times down for horizontal
\(F\), or \(S\) times left and \(S\) times right for vertical \(F\).
For each of the resulting placements, the brightness of subsequent pixels
in the upsampled
image, covered by the moved \(F\),
is determining the brightness functions \(B^{b}_{i}(x)\),
\(x = 0,\, \ldots\, N_{B} - 1\), where
\(N_{B}\) is the number of pixels in \(F\) and
\(i = -S_{f}, -S_{f} + 1,\, \ldots\, S_{f}\) is assigned
for each move of \(F\) as shown in the example in
Fig.~\ref{fig:frequency-filtering}, The index \(b = 0, 1,\, \ldots\, c - 1\)
determines one of the \(c\) bands of the upsampled image.
Each of these functions is subject to frequency
filtering as described in the next section.
\psfigure{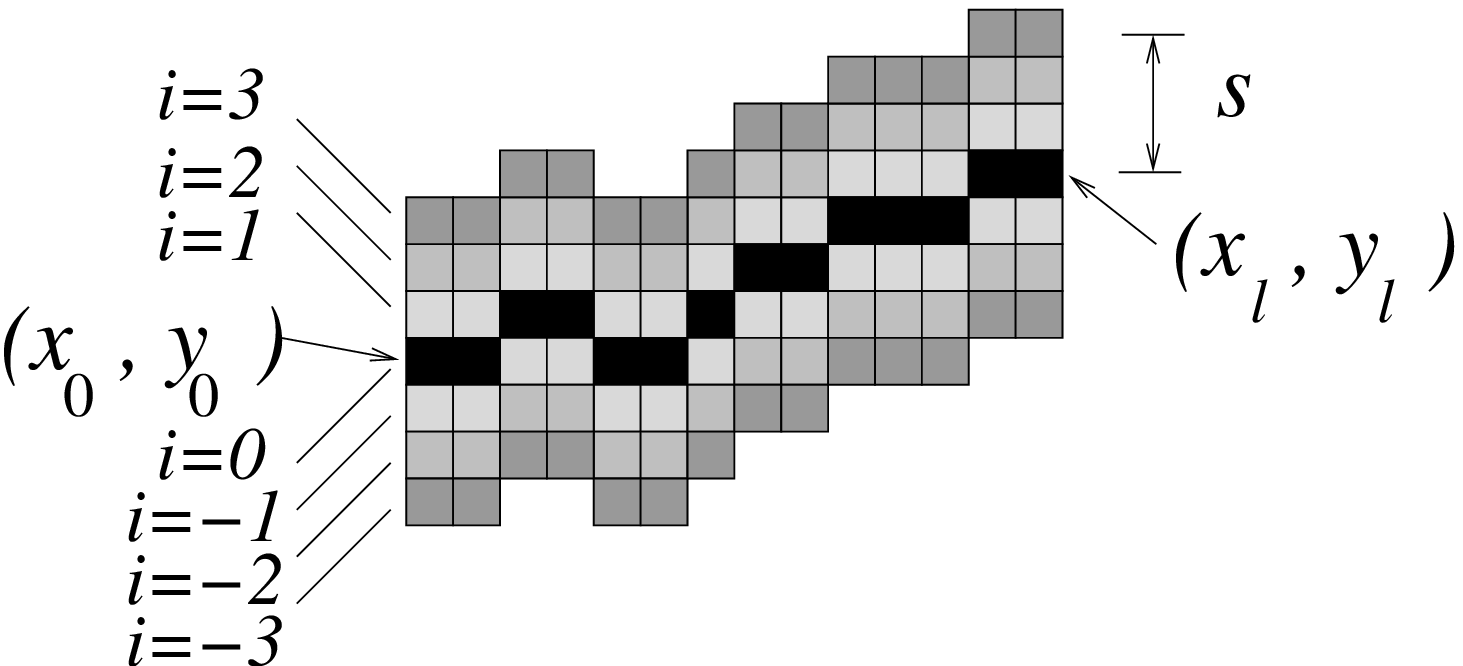}{2.3in}
{An example of a region filtered along a fragment. The fragment is
marked black.}

The pixels across different \(i\) do not overlap,
that is the band within each pixel is frequency filtered once, thanks to
the condition \(Q\) described in
Sec.~\ref{sec:approximately-straight-fragments}.

The variability of \(S_{f}\) comes from the
presume that an aliasing artifact is better detectable if it has the size of
at least several lengths of \(l_{0}\).
It is because the artifact might be otherwise
too easily mistaken with something that
is not such an artifact. For example a region being a normal
image matter without any artifacts
might likely have the brightness that is approximately given by a fragment
of a single lobe of the sine function that has the period of \(l_{0}\), yet
it might be much less likely that the brightness of that region is approximately
given by as much as several lobes of such a sine function.

The formula
for computing \(S_{f}\) on basis on the number of pixels \(N_{B}\)
in \(F\) is as follows:
\begin{equation}
	S_{f} = \left\{
		\begin{array}{ll}\displaystyle
		0 &
			\textrm{if \(N_{B} < s_{l}l_{0}\)} \\
		\displaystyle
		s_{u}N_{B} &
			\textrm{if \(N_{B} \ge s_{l}l_{0}\)} \\
		\end{array}
	\right.
\end{equation}
As it can be seen, \(S_{f} = 0\) if the fragment is too short, to decrease the
probability of confusing an artifact with image matter, as discussed earlier
in this section. Otherwise, \(S_{f}\) gradually increases with \(s_{u}N_{B}\). The
coefficients \(s_{l}\) and \(s_{u}\) were tuned in a series of tests. Small
\(s_{l}\) means a greater probability of an undesired distortion caused by
the frequency filtering of image
objects that are not artifacts. Conversely, large \(s_{l}\) means that
more artifacts might be left uncorrected. The coefficient \(s_{u}\)
regulates the strength of \(S\), which in turn is connected with the range
along \(F\) that is filtered. Thus, small \(s_{u}\) means that an
artifact might be corrected only in its small part closer to \(F\), and
large \(s_{u}\) means that some regions lying further to \(F\) might
be undesirably distorted by the filtering. It was found experimentally, that
\(s_{l} = 2\) and  \(s_{u} = 0.25\) give relatively good results.

\subsection{Filtering of the brightness function}
\begin{figure}[h!]
        \begin{center}
	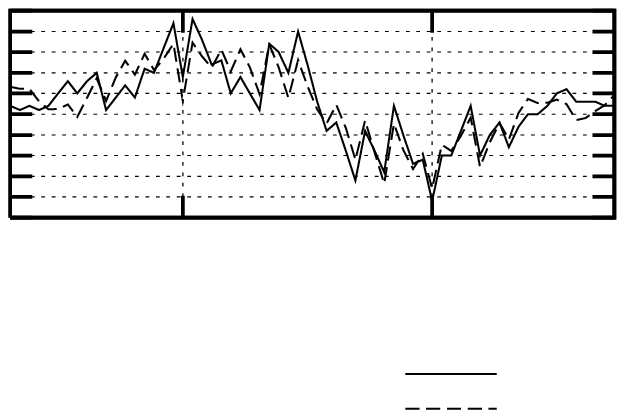
        \end{center}
	\vspace{-0.2in}
        \caption{An example of the functions \(C^{b}_{i}(x)\) and
         \(C'^{b}_{i}(x)\). Their fragments from
	 index 18 to index 44 are exact repetitions of, respectively, 
	 \(B^{b}_{i}(x - 18)\) and \(B'^{b}_{i}(x - 18)\).}
        \label{fig:chart-buffer}
        \end{figure}

The FFT requires the transformed function to have the number of
pairs to be the power of 2, what is untrue in general for
\(B^{b}_{i}(x)\). To prevent spurious high frequency
components and to fulfill the requirement for the number of pairs,
\(B^{b}_{i}(x)\) is padded with additional elements to create \(C^{b}_{i}(x)\).
Let the number of pairs
in \(C^{b}_{i}(x)\) be such a smallest possible value \(N_{C}\) that
it is the power of 2 and
the number of pad pairs \(N_{C} - N_{B}\)
to add to \(B^{b}_{i}(x)\) to create \(C^{b}_{i}(x)\)
is greater than or equal to \(\lfloor N_{B}/2 \rfloor\), Let the
mean value of \(B^{b}_{i}(x)\) be \(t^{b}_{i}\). The function
\(C^{b}_{i}(x)\) is defined as follows:
\begin{equation}
\label{eq:padding}
\begin{array}{c}
	x = 0 \ldots N_{C} - 1 \\
	m_{C} = \lfloor (N_{C} - N_{B})/2 \rfloor \\
	e_{C} = \lfloor m_{C} + N_{B} - 1 \rfloor \\
	w_{l} = x/(m_{C} - 1) \\
	w_{r} = (N_{C} - 1 - x)/(N_{C} - 2 - e_{C}) \\[8pt]
	C^{b}_{i}(x) = \left\{ \begin{array}{ll}
		w_{l}B^{b}_{i}(m_{C} - x) + (1 - w_{l})t^{b}_{i} & \textrm{for \(x < m_{C}\)} \\
		B^{b}_{i}(x - m_{C}) & \textrm{for \(x \ge m_{C} \,\land\)} \\
		& \textrm{\(x \le e_{C}\)} \\
		w_{r}B^{b}_{i}(2e_{C} - x) + (1 - w_{r})t^{b}_{i} & \textrm{for \(x > e_{C}\)} \\
	\end{array}\right.
	
\end{array}
\end{equation}
Thus, there are two mirror margins added with their widths
of at least \(\lfloor N_{B}/4 \rfloor\) each, that converge to the mean value of
\(B^{b}_{i}(x)\) at the lowest and the highest arguments of \(C^{b}_{i}(x)\).
The requirement for minimum width of the margins and the common convergence
value minimize spurious high frequency components in the spectrum of
\(C^{b}_{i}(x)\).
There
is an example of the function \(C^{b}_{i}(x)\) in Fig.~\ref{fig:chart-buffer}.

\begin{figure}[h!]
        \begin{center}
	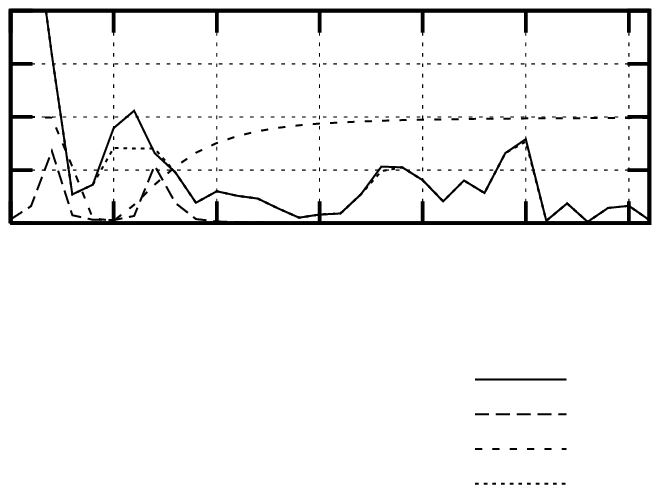
        \end{center}
	\vspace{-0.2in}
        \caption{An example of filtering of a function \(B^{b}_{i}(x)\) from
         Fig.~\ref{fig:chart-buffer}. The functions
         \(W(f)\) and \(M(f)\) are so
         distorted because of the low value of \(N_{C}\).}
        \label{fig:chart-frequency-filtering}
        \end{figure}

Let the brightness function \(C^{b}_{i}(x)\) after transforming it with
the FFT
be \(F^{b}_{i}(f)\), \(f = 0, \ldots\, N_{C} - 1\). Because  \(C^{b}_{i}(x)\)
is real, it holds true that \(F^{b}_{i}(f) = F^{b}_{i}(N_{C} - 1 - f)\)
for the whole domain of \(F^{b}_{i}(f)\). Further, because of that symmetry,
each operation to \(F^{b}_{i}(f)\) will also implicitly be applied to
\(F^{b}_{i}(N_{C} - 1 - f)\), and charts will show \(F^{b}_{i}(f)\)
only for \(f = 0, \ldots\, N_{C}/2 - 1\).
Let \(f_{0} = N_{C}/l_{0}\) be the frequency corresponding to \(l_{0}\).
If \(B^{b}_{i}(x)\) contains aliasing, peaks are expected near
to \(f_{0}\) and its harmonics in \(F^{b}_{i}(f)\). Because altering only the
peak at \(f_{0}\) appeared to be very effective, the peaks
at harmonics of \(f_{0}\) are ignored.
Simply setting the values of \(F^{b}_{i}(f)\) at or near \(f_{0}\) to 0
might produce the valley that would distort fragments that do not have
any artifacts, because a valley might appear in the frequency spectrum
where there was not even
any peak resulting from the aliasing. The solution is to compute the mean \(m\)
around the expected
peak at \(f_{0}\), and if the modulo of peak values exceed \(m\),
flatten the
peak down to the mean \(m\). The mean \(m\) is weighted using the weight function
\(W(f)\) such that it
has its maximum values at approximately \(1/2\) and \(3/2\) of
\(f_{0}\), and thus, these regions of maximum values
are placed away from both \(f_{0}\) and the harmonics of \(f_{0}\),
which in turn could contain peaks resulting from the aliasing,
and thus skew the value of \(m\).
The corresponding formulas for computing \(m\) are as follows:
\begin{equation}
\begin{array}{c}\displaystyle
\begin{array}{rl}\displaystyle
	W(f) =  &
		1/\big[1 + w_{s}(f - \frac{1}{2}f_{0})^{2}\big] + \\[6pt]
	& 1/\big[1 + w_{s}(f - \frac{3}{2}f_{0})^{2}\big] \\[6pt]
\end{array} \\
	\displaystyle
	S = \sum_{f = 0}^{f < N_{C}}W(f) \\[16pt]
	\displaystyle
	m = \frac{\sum_{f = 0}^{f < N_{C}}W(f)F^{b}_{i}(f)}{S}
\end{array}
\end{equation}
The coefficient \(w_{s}\) determines the width of each of
the two peaks and was tuned to 3 using test images.
The value \(S\) is computed to normalize the weight function \(W(f)\).
An example diagram of \(W(f)\) is shown in
Fig.~\ref{fig:chart-frequency-filtering}.

The reduction of the peak is in detail performed as follows.
Firstly, let the peak be located by a function \(M(f)\).
The value of the function is interpreted as follows: \(1\) for
no altering of the spectrum at \(f\), \(0\) for
a maximum altering of the frequency spectrum at \(f\), that is,
lowering its modulo values
to \(m\) if greater, Values of \(M(f)\)
in between \(0\) and \(1\)
determine respective partial alteration. The function
\(M(f)\) is computed as follows:
\begin{equation}
M(f) = \left\{\begin{array}{ll}
	1 &
	\textrm{if \(f = 0\)} \\
	\tanh\left[m_{s}\left(N_{C}/f - l_{0}\right)^2\right] &
	\textrm{if \(f > 0\)} \\
       \end{array}\right.
\end{equation}
The function is constructed so that \(M(f)\) creates a valley
at and near the peak with the lowest value close to \(0\),
is equal to \(1\) at the
constant component \(f = 0\), almost equal to \(1\) for frequencies
substantially lower or higher than \(f\).
The coefficient \(m_{s} = 0.03\)
was tuned to regulate the width and slopes of the valley.
The limited steepness of the slopes of \(M(f)\) reduces the the
possible distortions in the space domain caused by the frequency filtering.
The left slope is so steep to decrease the reduction of low frequencies.
Reducing them, because of their usually large values,
appeared to produce strong discontinuity effects between
the filtered region and the rest of the image.

The discussed alteration of
\(F^{b}_{i}(f)\) so that it creates \(F'^{b}_{i}(f)\) has the following
equation:
\begin{equation}
\forall_{f}\quad
|F'^{b}_{i}(f)| = \left\{\begin{array}{ll}
	 M(f)|F^{b}_{i}(f)| + & \\
         \quad + \big[1 - M(f)\big]m  & \textrm{if \(|F^{b}_{i}(f)| > m\)} \\
	 |F^{b}_{i}(f)| & \textrm{if \(|F^{b}_{i}(f)| \le m\)} \\
       \end{array}\right.
\end{equation}
An example of filtering of \(F^{b}_{i}(f)\)
is shown in Fig.~\ref{fig:chart-frequency-filtering}.
The function \(F'^{b}_{i}(f)\) is transformed using reverse FFT
into \(C'^{b}_{i}(x)\), from which is extracted
\(B'^{b}_{i}(x) = C'^{b}_{i}(x - m_{C})\), \(x = 0\, \ldots\, N_{B} - 1\),
to remove the padding introduced in (\ref{eq:padding}).
\(B'^{b}_{i}(x)\) is thus a frequency--filtered \(B^{b}_{i}(x)\), and is written back
to the upsampled image, to the exact pixels from which \(B^{b}_{i}(x)\)
was constructed.

\section{Tests}
\label{sec:tests}
An example image processed with the presented method is shown in
Fig.~\ref{fig:edge-detection}(d). The image was upsampled four times
using bicubic interpolation that employed Catmull--Rom spline
\cite{mitchell1988reconstruction}. The edge map of the image is
shown in Fig.~\ref{fig:edge-detection}(b). As it can be seen, the image
with reduced aliasing is visually
radically improved over the image obtained using plain upscaling without
the frequency filtering, shown in Fig.~\ref{fig:edge-detection}(c).
The aliasing in Fig.~\ref{fig:edge-detection}(d) is almost
reduced, without any substantial blur, loss of small details
or other distortions visible. It differs the presented method from that of
an anisotropic smoothing \cite{tschumperle2006anisotropic}
shown in Fig.~\ref{fig:flower_128_greyc}, which, while reducing
the aliasing, distorts the image so that it looks very unnatural
and blurred. For example, most of the details in the center of the petal
in Fig.~\ref{fig:flower_128_greyc} are almost lost.
		\psfigureabcd{edge-detection}
		{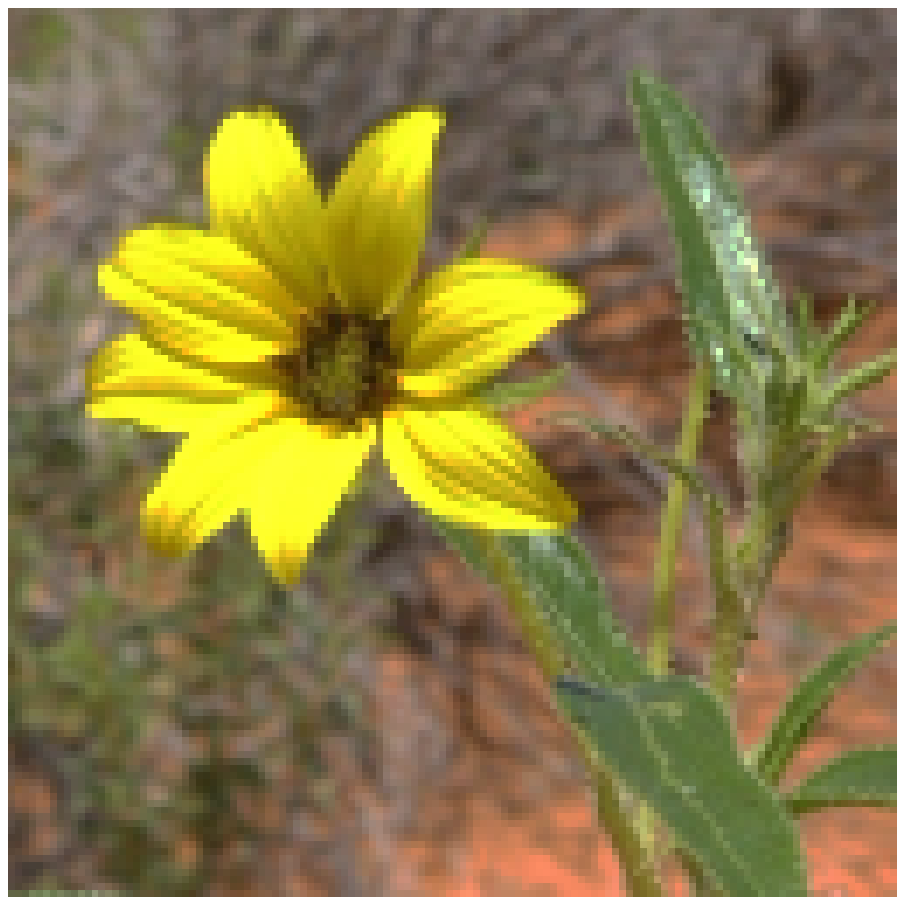}
		{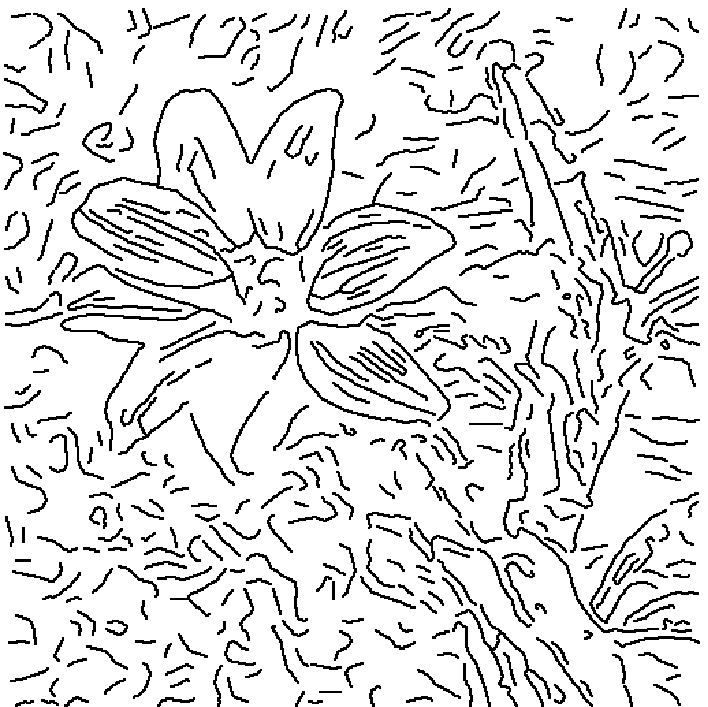}
		{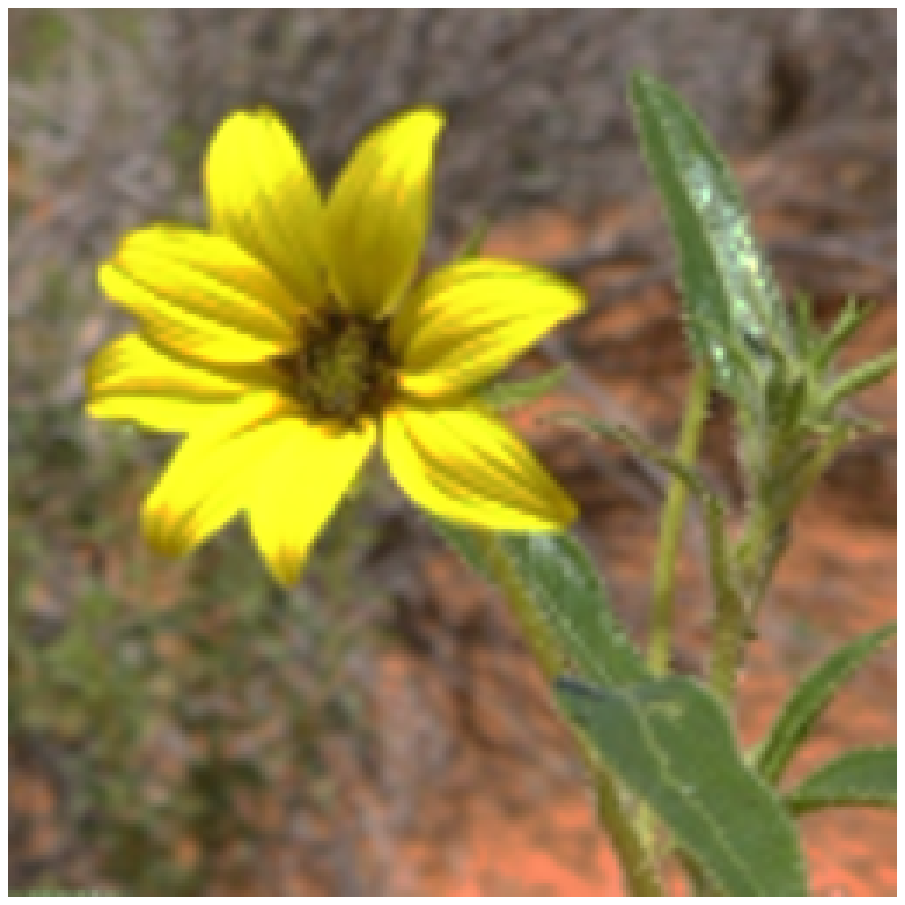}
		{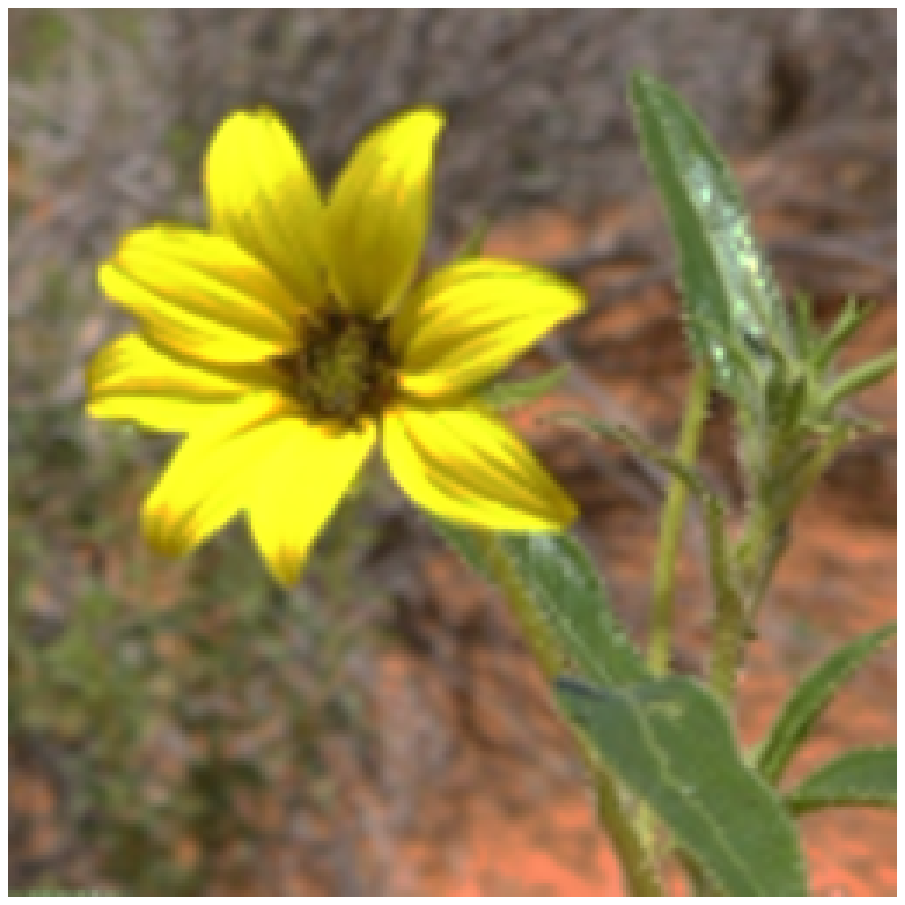}
		{1.8in}
                {An example of filtering a photograph: (a) original image,
                 (b) subpixel precision edges found, thickened in the
                 illustration to make them better visible,
                 (c) upsampled image without the frequency filtering,
                 (d) upsampled image with the frequency filtering.}
\psfigure{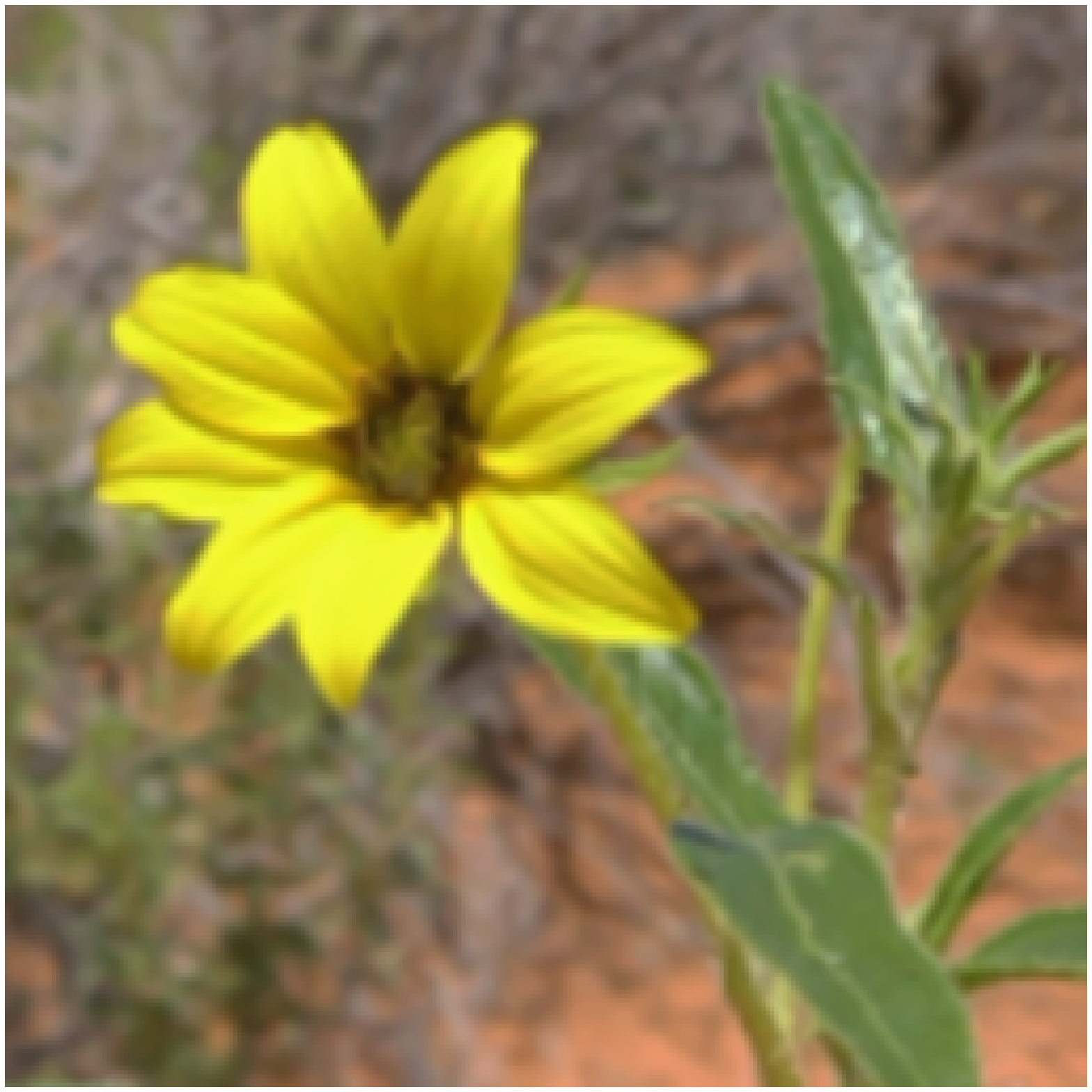}{1.8in}
{The image from Fig.~\ref{fig:edge-detection}(a)
upsampled using a GREYC anisotropic smoothing.}

It can be seen in the image, that the introduced
method works well for various non--straight curves,
even that it splits them into the approximately straight fragments
before the frequency filtering.

\section{Conclusion}
The presented method can be applied to images upsampled using
different interpolation method, and can radically reduce
aliasing, with a very good preservation of the rest of the
filtered image. The method has the side effect of producing
a subpixel precision edge map, that can be used in various
edge processing algorithms, like the sharpening of edges
in the upsampled image, for further improvement of its quality.

{\scriptsize
\bibliography{ip}
\bibliographystyle{plain}\setlength{\itemsep}{-4mm}
}

\end{document}